# Dynamic Object Detection and Tracking in Construction: A Fisheye Camera and LiDAR Sensor Fusion Model

Yilong Chen, Huili Huang, and Yong K. Cho

***Abstract*-- Robust dynamic object detection and tracking are essential for enabling robots to operate safely and effectively alongside humans in complex environments such as construction sites. While LiDAR-based SLAM and occupancy grid methods offer viable solutions for detecting and tracking motion, many state-of-the-art 3D vision approaches rely heavily on pre-trained neural networks and require additional post-processing to identify moving objects. Sensor fusion techniques, combining the precision of LiDAR with the semantic richness of RGB imagery, offer a promising alternative. In this work, we present a novel framework that enhances a quadruped robot equipped with a LiDAR sensor and an upward-facing fisheye camera for real-time dynamic object detection and tracking. After identifying moving objects within a registered point cloud, our method assigns semantic labels by projecting 3D coordinates onto a 2D cylindrical panorama, aligning with real-time image-based detections for observation update of the Kalman filter. The proposed system demonstrates high precision, simplicity, and robustness, particularly in handling objects transitioning between dynamic and static states, thus it is well-suited for deployment in real-world construction environments.**

## I. Introduction

Simultaneous Localization and Mapping (SLAM) has been widely adopted for legged robots equipped with onboard sensors to reconstruct both indoor [1] and outdoor [2] environments, particularly in response to the growing integration of robotic systems in construction tasks. However, dynamic objects are often treated as noise in the SLAM problem, leading most existing approaches to separate dynamic object tracking from map reconstruction [3]. Despite this, discrepancies between consecutive point cloud maps can offer critical insights into object motion relative to the sensor, presenting an opportunity to integrate perception and mapping more effectively. In prior work [4], we proposed an online dynamic object detection and tracking method based on LiDAR SLAM and occupancy grids. By assigning discrete states to grid cells and updating occupancy probabilities through discounted returns across state transitions, dynamic objects can be clustered. This information is then fed into a Kalman filter for tracking and motion state estimation, without requiring any additional sensing beyond the LiDAR.

Nevertheless, two key challenges persist: (1) the inability to semantically identify detected moving objects, and (2) degraded tracking performance when objects transition from dynamic to temporarily static states. While incorporating a camera can address these issues by providing semantic context, existing deep learning-based 3D object detection approaches often demand complex models, substantial computational resources, extensive labeled datasets, and heavy

YC, HH, YC are with Georgia Institute of Technology, Atlanta, GA 30332, USA, {ychen3339, hhuang413}@gatech.edu, yong.cho@ce.gatech.edu

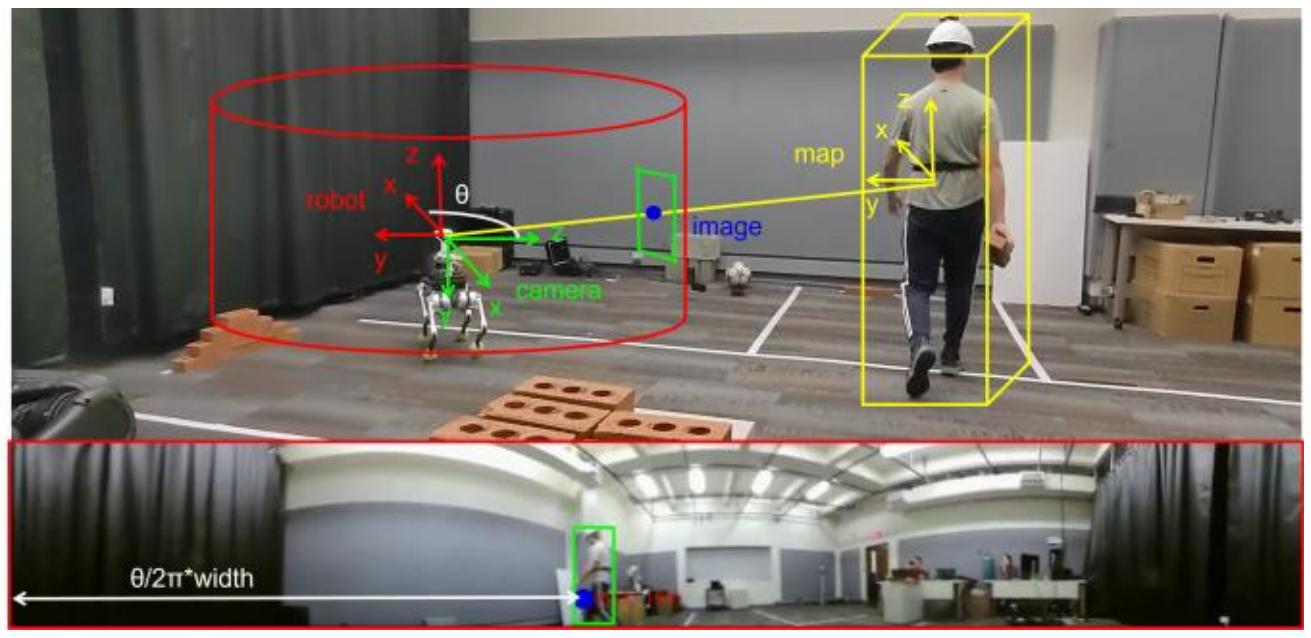

Figure 1. Frame System (top) and Point Projection and Mapping (bottom).

dependence on accurate depth estimation [5], which can be error-prone. Conventional RGB cameras also suffer from limited fields of view, and using multiple cameras introduces further challenges in extrinsic and temporal calibration [6]. Additionally, distortion correction is required when using pinhole lens models. To address these limitations, we integrate a single upward-facing fisheye camera into our existing LiDAR-based system, enabling a unified and semantically rich perception framework.

This research proposes a novel method for dynamic object detection and tracking by combining registered LiDAR point cloud maps from SLAM with cylindrical panoramic images derived from the fisheye camera (Fig. 1). The framework comprises two core modules: (1) a projection and labeling pipeline that maps 3D dynamic objects onto a 2D cylindrical panorama for semantic annotation, and (2) a dynamic object tracking mechanism based on occupancy grid updates and a modified Kalman filter. We validate our approach through experiments conducted on a quadruped robot operating in a real-world indoor construction setting. The primary contributions of this work are as follows: (a) the development of a generalizable LiDAR-camera fusion method for dynamic object detection and tracking to support robotic navigation, (b) enhanced detection and tracking accuracy through semantic augmentation via sensor fusion, and (c) successful deployment and validation of the proposed system in a real-world construction environment.

## II. Related work

With the rise of autonomous systems, such as autonomous robots and on-road self-driving vehicles, sensor fusion is an essential technology that can lessen detection uncertainty and overcome individual sensors' drawbacks while working alone. For example, integrating LiDAR and the camera could provide semantically rich images along with accurate depth measurement and velocity estimation. Utilizing the detailed lane geometry provided by Argoverse's high-definition maps, [7] expands 2D instance segmentations from image space into 3D cuboids within LiDAR space, thereby mitigating ambiguities in object orientation. These self-annotated, inflated cuboids serve as effective training targets for 3D object detection models. In contrast, [8] project LiDAR point clouds onto the image plane and estimate object depth by averaging the calibrated LiDAR points that fall within the 2D bounding boxes. Furthermore, advancements in attention mechanisms have facilitated the integration of multiple transformer modules for cross-modal feature fusion. Leveraging this, [9] introduces a Multi-Modal Fusion Transformer that processes LiDAR bird's-eye view (BEV) representations alongside single-view RGB images, enabling robust perception in complex urban driving environments.

Unlike the former approaches that need tremendous labeled point cloud data, accurate calibration between a LiDAR sensor and camera, or extensive computation resources, our methods acquire and utilize the label from a YOLOv8 [10] detector in a bi-directional way. On the one hand, we only project detected dynamic object centers onto a cylindrical panorama view generated by a 360° fisheye camera set to face upward; thus, the horizontal angle θ, or azimuth, to the detected dynamic object following image transformation is unaffected by the camera distortion. On the other hand, the mapping process between detection results from the 2D images and the 3D point cloud could reduce both the false positive objects originating from the pretrained model in the image, as well as false negative temporary static objects caused by the loose tracking issue in the point cloud (Fig. 2).

## III. Method

### A. LiDAR-Camera Fusion and Projection

Following the identification of dynamic objects in the environment, semantic labels are assigned through sensor fusion between LiDAR and a camera system. A 360° fisheye camera mounted on the quadruped robot provides panoramic semantic information in a cylindrical projection view. As shown in Fig. 1, the centers of the detected dynamic objects are first transformed from the global map frame to the robot's local coordinate frame using pose estimates obtained from the SLAM module, and subsequently to the camera coordinate frame. These 3D centers are then projected onto the cylindrical panoramic canvas to spatially align with 2D object labels detected in real time by the YOLOv8 detector. Leveraging its strong generalization capability, we implement a zero-shot YOLOv8 model to identify workers without requiring additional task-specific fine-tuning.

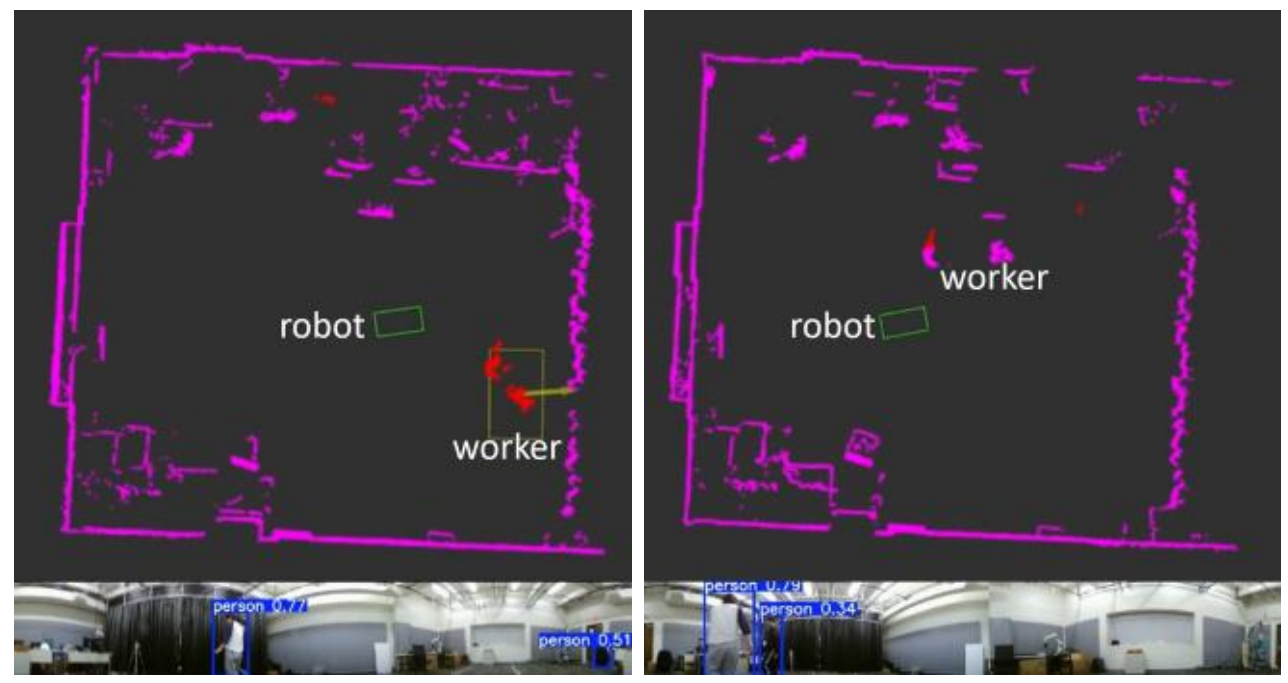

Figure 2. False Positive Objects in the Image (left) and False Negative Temporary Static Objects in the Point Cloud (right).

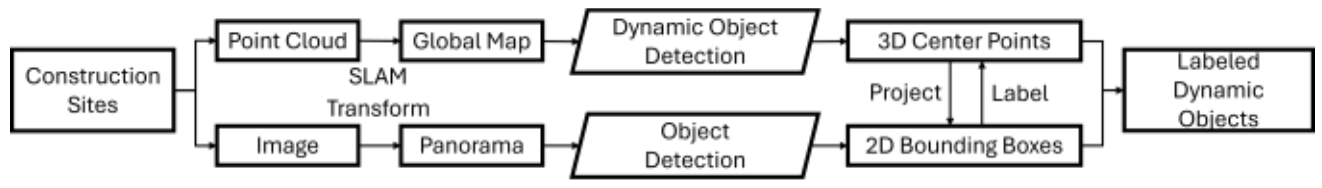

Figure 3. Workflow of the LiDAR-Camera Fusion Model

Due to the upward-facing configuration of the fisheye camera, image distortion does not affect the horizontal angle θ after projection. This azimuth angle is used to match the projected center point of each dynamic object with the corresponding 2D bounding box. Once matched, the semantic label from the image is mapped back to the corresponding 3D object. The complete workflow of this fusion and projection process is illustrated in Fig. 3.

### B. Dynamic Object Detection

We proposed a dynamic object detection method in [4], integrating mapping and robot pose data from SLAM for the recognition of dynamic objects. Compared with previous occupancy grid-based methods, we made several improvements in the following aspects. First, instead of modeling the whole observing space or transition probability, we just update the state value function for the target grid as needed. Second, in addition to the free and occupied states, we define the extra unobserved and neighbor states to improve the responsiveness to dynamic changes on construction sites. Third, we incorporate memory weight and discounted reward to balance the immediate and long-term impacts of state transitions on state values. Finally, we are able to achieve online predictions by incrementally updating the occupancy probability rather than using a neural network. $\bar{G}_{t+1}$, $\bar{G}_t$, $\bar{R}_{t+1}$, $\gamma$ in formula (1) are normalized total return in frame $t+1$ and $t$, normalized reward for state transition from frame $t$ to $t+1$, and discounted rate, respectively.

$$\bar{G}_{t+1} = \bar{R}_{t+1} + \gamma * \bar{G}_t \tag{1}$$

Fig. 4 illustrates the process of determining the states and the reward between state transitions. The order in which we identify the cell states in practice is unobserved, occupied, free, and neighbor. First, we make every cell on the global map unobserved. We then iterate through each point and set the relevant cells to be occupied. Then we perform ray tracing from the LiDAR sensor to each of the occupied cell's center. All cells on the route to the occupied cell are set as free. Finally, we designated all eight cells around a cell as neighbors unless they were already set occupied. We simply set the rewards to $-R$, $R$, 0, and $R$ for state transitions from free, occupied, unobserved, and neighbor states, respectively, which gather the rewards from the past to enable online learning. Readers are recommended to go through [4] for further details and experiment results.

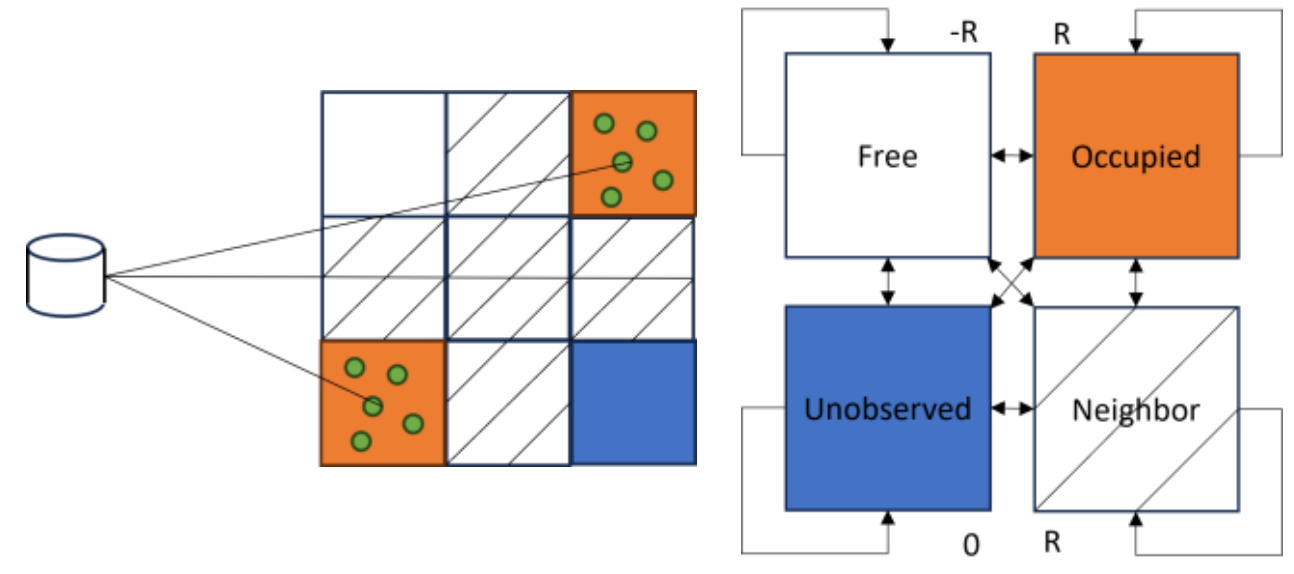

Figure 4. State Determination (left) and State Transition (right).

## C. Multiple Object Tracking

Building on our previous work [4], we adopt a multi-object association and tracking framework based on a modified Kalman filter, designed to handle edge cases where the number of detected objects varies between consecutive frames. In addition to tracking the object centroid, each Kalman filter maintains the half-width of the bounding box to preserve size information. A "life" parameter is also maintained to manage the persistence of tracked objects across frames.

Following the estimation of the centroids and half widths of the dynamic objects in a single frame, we proceed to track the motion state of the dynamic objects using Kalman filters. The transition matrix $F$, observation matrix $H$, measurement vector $z_k$, and state vector $x_k$ can all be expressed as (2-5)

$$x_k = [x_{c,k}, y_{c,k}, v_{x,k}, v_{y,k}]^T \tag{2}$$

$$z_k = [x_{c,k}, y_{c,k}]^T \tag{3}$$

$$F = \begin{bmatrix} 1 & 0 & dt & 0 \\ 0 & 1 & 0 & dt \\ 0 & 0 & 1 & 0 \\ 0 & 0 & 0 & 1 \end{bmatrix} \tag{4}$$

$$H = \begin{bmatrix} 1 & 0 & 0 & 0 \\ 0 & 1 & 0 & 0 \end{bmatrix} \tag{5}$$

While $v_{x,k}, v_{y,k}$ represent the traveling speeds in the $x$ and $y$ directions in frame k, respectively, $x_{c,k}, y_{c,k}$ represent the center coordinates of the bounding box in frame $k$. The standard Kalman filter updates its state vector whenever a new measurement becomes available.

In order to solve the problem of different number of objects between two consecutive frames, we maintain two obejcts arrays: the newly detected dynamic objects *Objs* with under-tracking objects *TrackingObjs*. The basic idea is to associate newly detected objects with under-tracking objects in a certain distance threshold. The "life" parameter gets reset to full if it is associated with a newly detected object; otherwise it will be deducted. Details of the algorithm can be found in [4] .

A common challenge arises when an object becomes temporarily static, its "life" value may rapidly decay to zero, resulting in premature disappearance and degraded tracking performance. To address this problem, we increase the “life” value whenever a dynamic object is confirmed by the image-based detector’s bounding box predictions. For static objects not matched with any detections, we bypass the motion prediction step but still perform the observation update using the last known state. This allows the estimated velocity to converge toward zero, preserving tracking continuity and accuracy for stationary objects.

# IV. Experiment for Lidar-camera sensor fusion

## A. Experiment Environment and Setting

To evaluate the overall performance of the proposed method in dynamic object detection and tracking under real-world conditions, we designed an indoor bricklaying scenario, illustrated in Fig. 5. The experimental environment consisted of two sets of ten bricks positioned at both the long and short ends of a cross-shaped corridor. Two participants were instructed to separately construct brick walls at the end of the longer corridor and the shorter corridor, respectively. During the task, a quadruped robot (shown in Fig. 6) was guided alongside the participant along the longer corridor using an autonomous navigation package.

To obtain ground-truth position data for both the participant and the robot, we employed a Vicon motion capture system with reflective markers attached to each. The design of the environment layout and the human-robot interaction protocol draws upon our prior research in this domain [11], enabling the robot to experience crossing, passing, merging, and group interactions with human in a controlled setting.

For spatial representation, we utilized a 35×35 occupancy grid with a resolution of 0.3 meters per cell. The SLAM system was implemented using the LOL-SAM algorithm [12]. A LiDAR sensor captured environmental data at 10 Hz with 1600 samples per scan. The dynamic object detection module processed the raw, registered point clouds published at 10 Hz, while bounding boxes were generated in real-time using the YOLOv8 detector, also at 10 Hz. The total length of the recorded message topic streams is 150 seconds.

## B. Experiment Results

Fig. 7 presents the distribution of the estimated positions of the human participant, alongside the ground truth trajectories for both the human and the quadruped robot. Both the *LiDAR Only* and *Sensor Fusion* configurations demonstrate a strong alignment with the ground truth trajectories. However, a notable difference is observed at the endpoints of both corridors, where the *Sensor Fusion* approach yields a significantly higher density of estimated human positions. Specifically, the *Sensor Fusion* setup produced 2,061 position estimates—nearly double the 1,038 estimates generated by the *LiDAR Only* configuration. This outcome suggests that integrating LiDAR with a fisheye camera substantially improves the system’s ability to continuously track human workers during periods of temporary inactivity, such as while performing bricklaying tasks.

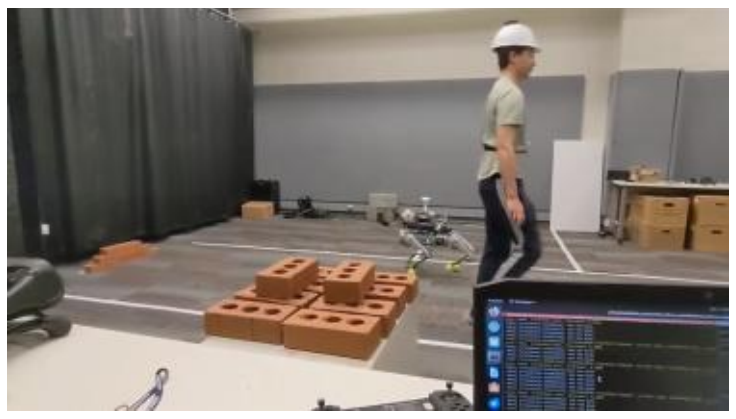

Figure 5. Bricklaying Experiment

Figure 6. Quadruped Robot

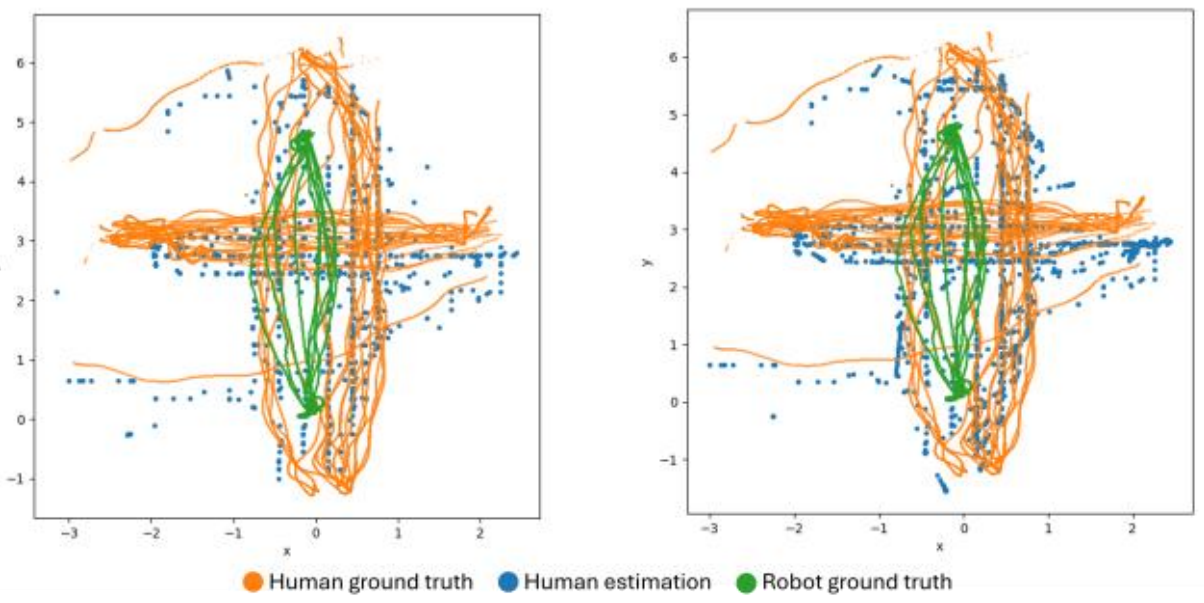

Figure 7. Human Detection Distribution. LiDAR Only (left) and Sensor Fusion (right).

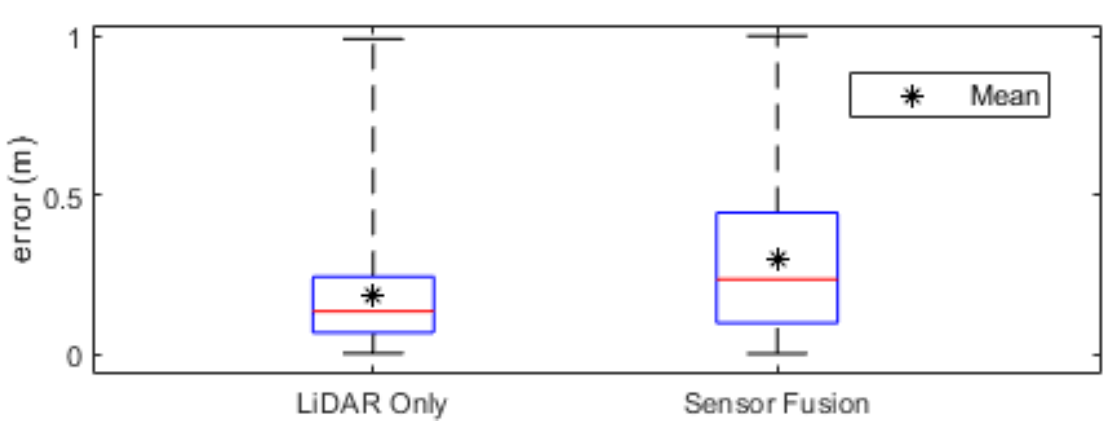

Figure 8. Human Distance Estimation Error. LiDAR Only (left) and Sensor Fusion (right).

To evaluate tracking accuracy, we compared the estimated distance between the participant and the robot against ground truth data captured by a Vicon motion capture system. Due to the Vicon system's higher sampling rate (approximately 12 times that of the registered point cloud) and occasional data loss, only frames with available ground truth were considered. Fig. 8 illustrates the human distance estimation errors for both configurations. Given the grid resolution of 0.3 meters used in our experimental setup, the average errors were 0.18 meters for the *LiDAR Only* system and 0.3 meters for the *Sensor Fusion* system. The slightly higher error in the *Sensor Fusion* results can be attributed to the increased number of estimates near the corridor endpoints. During transitions between dynamic and temporarily static states, our method halts position updates for static objects, whereas minor movements of reflective markers attached to the participant continue to be recorded, contributing to small discrepancies.

## V. Conclusion

This study introduces a novel sensor fusion framework for real-time dynamic object detection and tracking, leveraging the integration of registered LiDAR point cloud maps from SLAM with cylindrical panoramic views captured by a fisheye camera. Through experiments conducted in a real-world indoor construction environment, our approach demonstrates strong potential for enhancing robotic perception and navigation in complex, human-centric settings. The ability to track dynamic objects—even during transitions to temporary static states—highlights the effectiveness of our method in practical applications.

Looking ahead, future work will focus on utilizing the labeled positions of dynamic objects to support a more comprehensive understanding of the surrounding environment, ultimately enabling the development of socially aware navigation strategies.

Nonetheless, several challenges remain. The current projection of 3D point clouds into 2D space, based on occupancy probability changes, is limited when dealing with elevation variations or multi-level obstacles, necessitating more advanced 3D handling. Additionally, because our current approach matches dynamic objects based on proximity to the candidate object, incorrect associations between multiple object pairs may occur, highlighting the need for a more sophisticated matching algorithm to improve pairing accuracy. Finally, although the study emphasizes its advancements in lowering LiDAR false negatives and image false positives, it still lacks a statistical and comprehensive analysis of the effect and trade-off of both sensors.